# Learning From Labeled And Unlabeled Data: An Empirical Study Across Techniques And Domains

**Nitesh V. Chawla**                                             NCHAWLA@CSE.ND.EDU
*Department of Computer Science & Engg.,*
*University of Notre Dame,*
*IN 46556, USA*

**Grigoris Karakoulas**                                          GRIGORIS@CS.TORONTO.EDU
*Department of Computer Science*
*University of Toronto*
*Toronto, Ontario*
*Canada M5S 1A4*

## Abstract

There has been increased interest in devising learning techniques that combine unlabeled data with labeled data – i.e. semi-supervised learning. However, to the best of our knowledge, no study has been performed across various techniques and different types and amounts of labeled and unlabeled data. Moreover, most of the published work on semi-supervised learning techniques assumes that the labeled and unlabeled data come from the same distribution. It is possible for the labeling process to be associated with a selection bias such that the distributions of data points in the labeled and unlabeled sets are different. Not correcting for such bias can result in biased function approximation with potentially poor performance. In this paper, we present an empirical study of various semi-supervised learning techniques on a variety of datasets. We attempt to answer various questions such as the effect of independence or relevance amongst features, the effect of the size of the labeled and unlabeled sets and the effect of noise. We also investigate the impact of sample-selection bias on the semi-supervised learning techniques under study and implement a bivariate probit technique particularly designed to correct for such bias.

## 1. Introduction

The availability of vast amounts of data by applications has made imperative the need to combine unsupervised and supervised learning (Shahshahni & Landgrebe, 1994; Miller & Uyar, 1997; Nigam et al., 2000; Seeger, 2000; Ghani et al., 2003). This is because the cost of assigning labels to all the data can be expensive, and/or some of the data might not have any labels due to a selection bias. The applications include, but are not limited to, text classification, credit scoring, and fraud and intrusion detection. The underlying challenge is to formulate a learning task that uses both labeled and unlabeled data such that generalization of the learned model can be improved.

In recent years various techniques have been proposed for utilizing unlabeled data (Miller & Uyar, 1997; Blum & Mitchell, 1998; Goldman & Zhou, 2000; Nigam et al., 2000; Blum & Chawla, 2001; Bennett et al., 2002; Joachims, 2003). However, to the best of our knowledge, there is no empirical study evaluating different techniques across domains and data distributions. Our goal in this paper is to examine various scenarios of dealing with labeled and unlabeled data in conjunction with multiple learning techniques.

Incorporating unlabeled data might not always be useful, as the data characteristics and/or technique particulars might dictate that the labeled data are sufficient even in the presence of a corpus of unlabeled data. There have been conflicting experiences reported in the literature. On the one hand, the NIPS'01 competition on semi-supervised learning required the contestants to achieve better classification results in most of the cases than just supervised learning (Kremer & Stacey, 2001). This implies an expectation that in most of the cases semi-supervised techniques could help. However, the results from the competition did not meet this expectation. Shahshahni and Landgrebe (1994) note that unlabeled data can delay the occurrence of Hughes phenomenon and the classification performance can be improved. Zhang and Oles (2000), using the Fisher Information criterion, show that for parametric models unlabeled data should always help. On the other hand, Cozman et al. (2002; 2003), using asymptotic analysis, show that unlabeled data can in fact decrease classification accuracy. They concluded that if the modeling assumptions for labeled data were incorrect, then unlabeled data would increase classification error, assuming the labeled and unlabeled data come from the same distribution. But they did not formulate the asymptotic behavior of semi-supervised techniques on scenarios in which the labeled and unlabeled data come from different distributions. In addition, they only evaluate their algorithm on one real dataset, by varying the amount of labeled and unlabeled data.

The aforementioned semi-supervised learning techniques do not distinguish the different reasons for the data being missing. Most of the techniques assume that the data are *missing completely at random (MCAR)*, i.e. $P(labeled=1 | x, y) = P(labeled=1)$ where $y$ is the class label assigned and *labeled=1* if a given example is labeled (Little & Rubin, 1987). In this case the labeling process is not a function of variables occurring in the feature space. The labeled and unlabeled data are assumed to come from the same distribution.

When the labeling process is a function of the feature vector $x$ then the class label is *missing at random (MAR)*, i.e. $P(labeled=1 | x, y) = P(labeled=1|x)$. This scenario of MAR labeling can occur, for example, in credit scoring applications since many creditors use quantitative models for approving customers. In that case the class label $y$ is observed only if some (deterministic) function $g$ of variables in $x$ exceeds a threshold value, i.e. $g(x) \geq c$, where $c$ is some constant. From the definition of MAR it follows that $P(y=1 | x, labeled=1) = P(y=1 | x, labeled=0) = P(y=1|x)$, i.e. at any fixed value of $x$ the distribution of the observed $y$ is the same as the distribution of the missing $y$. However, due to the aforementioned screening, the conditional distribution of $x$ given $y$ is not the same in the labeled and unlabeled data, i.e. there is a bias. Thus, in modeling techniques that aim to learn such conditional distributions one needs to correct for this bias by incorporating information from unlabeled data.

In contrast, when labeling depends on $y$, the class label is *missing not at random (MNAR)*, i.e. $P(labeled=1 | x, y) \neq P(labeled=0 | x, y)$. In that case there is *sample-selection bias* in constructing the labeled set that has to be corrected (Heckman, 1979). For example, sample-selection bias can occur in credit scoring applications when the selection (approval) of some customers (labels) is performed based on human judgment rather than a deterministic model (Greene, 1998; Crook & Banasik, 2002; Feelders, 2000). Sample-selection bias can also occur in datasets from marketing

campaigns. In the MNAR scenario the labeled and unlabeled data come from different distributions associated with a censoring mechanism, i.e. $P(y=1|x, labeled=1) \neq P(y=1|x, labeled=0)$. In such a scenario learning only from labeled data can give estimates that are biased downwards. When a selection bias is associated with the labeling process it is necessary to model the underlying missing data mechanism. Shahshahani and Landgrebe (1994) also note that if the training sample is not very representative of the distribution in the population then the unlabeled data might help.

In this paper we present an empirical study of some of the existing techniques in learning from labeled and unlabeled data under the three different missing data mechanisms, i.e. MCAR, MAR and MNAR. To the best of our knowledge, the effect of unlabeled data under different missing data mechanisms has not been previously addressed in the semi-supervised learning literature. We also introduce two techniques from Econometrics, namely reweighting and bivariate probit, for semi-supervised learning. We try to answer various questions that can be important to learning from labeled and unlabeled datasets, such as:

- *What is the process by which data have become labeled vs. unlabeled?*
- *How many unlabeled vs. labeled examples in a dataset?*
- *What is the effect of label noise on semi-supervised learning techniques?*
- *Are there any characteristics of the feature space that can dictate successful combination of labeled and unlabeled data?*
- *What is the effect of sample-selection bias on semi-supervised learning methods?*

For our experiments we have chosen datasets with unbalanced class distributions, since they can be typical of real-world applications with unlabeled data, such as information filtering, credit scoring, customer marketing and drug design. For such datasets, accuracy can be a misleading metric as it treats both kinds of errors, false positives and false negatives, equally. Thus, we use AUC, Area Under the Receiver Operating Curve (ROC), as the performance metric (Swets, 1988; Bradley, 1987). AUC has become a popular performance measure for learning from unbalanced datasets (Provost & Fawcett, 2000; Chawla et al., 2002).

The paper is organized as follows. In Section 2 we present an overview of the techniques that were evaluated in this work. These techniques are mainly applicable to MCAR or MAR type of data. In Section 3 we present two techniques that purport to deal with sample-selection bias in MNAR data. In Section 4 we describe the experimental framework and in Section 5 we analyze the results from the experiments. We conclude the paper with a discussion of the main findings.

## 2. Semi-Supervised Learning Methods

The techniques we evaluate for learning from labeled and unlabeled data are: Co-training (Blum & Mitchell, 1998; Goldman & Zhou, 2000), Reweighting (Crook & Banasik, 2002), ASSEMBLE (Bennett et al., 2002) and Common-component mixture with EM (Ghahramani & Jordan, 1994; Miller & Uyar, 1997). We chose the variant of the co-training algorithm proposed by Goldman and Zhou (2000) since it does not make the assumption about redundant and independent views of data. For most of the datasets considered, we do not have a "natural" feature split that would offer redundant views. One could try to do a random feature split, but again that is not in-line with the original assumption of Blum and Mitchell (1998). Essentially, two different algorithms looking at same data can potentially provide better than random information to each other in the process of labeling. By construction, co-training and ASSEMBLE assume that the labeled and unlabeled data are sampled from the same distribution, namely they are based on the MCAR assumption. The re-weighting and common-component mixture techniques are based on the assumption that data are of MAR type. Thus, they can be applied to cases where the conditional distribution of $x$ given $y$ is not the same in the labeled and unlabeled data. Details of these techniques are given in the rest of this Section.

For sample-selection correction in MNAR data we use the Bivariate Probit technique (Heckman, 1979; Greene, 1998) and Sample-Select, an adapted version of the technique used by Zadrozny and Elkan (2000). Both of these techniques will be presented in Section 3.

It is worth pointing out that we selected some of the most popular algorithms for learning from labeled and unlabeled data. However, the list of algorithms under study is not meant to be exhaustive. This is because our main goal is to evaluate the behavior of such algorithms with different amounts, distributions, and characteristics of labeled and unlabeled data.

We used the Naïve Bayes algorithm as the underlying supervised learner for all the aforementioned semi-supervised learning methods. We chose Naïve Bayes due to its popularity in the semi-supervised learning framework as well as the need for a generative model for the common-component mixture technique. Thus, this choice for the base learner helped us achieve consistency and comparability in our experimental framework. Only for the co-training experiments, which by design require two classifiers, we also used C4.5 decision tree release 8 (Quinlan, 1992). Please note that in the rest of the paper, we will use Naïve Bayes and supervised learner, interchangeably.

We next introduce a notation that will be used for describing the various methods in this paper. Consider a classification problem with K classes $y_k$, $k = 0,1,...,K-1$. We will assume that the training set consists of two subsets: $X = \{X_L, X_U\}$, where

$X_L = \{(x_1, y_1), (x_2, y_2),...,(x_l, y_l)\}$ is the labeled subset,

$X_U = \{x_{l+1}, x_{l+2},...,x_{l+u}\}$ is the unlabeled subset,

$l$ and $u$ are the number of examples in the labeled and unlabeled sets, respectively, and

$x$ is the $n$-dimensional feature vector.

## 2.1 Co-training

Co-training, proposed by Blum and Mitchell (1998), assumes that there exist two independent and compatible feature sets or views of data. That is, each feature set (or view) defining a problem is independently sufficient for learning and classification purposes. For instance, a web page description can be partitioned into two feature subsets: words that exist on a web page and words that exist on the links leading to that web page. A classifier learned on each of those redundant feature subsets can be used to label data for the other and thus expand each other's training set. This should be more informative than assigning random labels to unlabeled data. However, in a real-world application, finding independent and redundant feature splits can be unrealistic, and this can lead to deterioration in performance (Nigam & Ghani, 2001).

Goldman and Zhou (2000) proposed a co-training strategy that does not assume feature independence and redundancy. Instead, they learn two different classifiers (using two different supervised learning algorithms) from a dataset. The idea behind this strategy is that since the two algorithms use different representations for their hypotheses they can learn two diverse models that can complement each other by labeling some unlabeled data and enlarging the training set of the other. Goldman and Zhou derive confidence intervals for deciding which unlabeled examples a classifier should label. We adopt this co-training strategy because it allows us to apply it to various real-world datasets and thus compare it with other semi-supervised learning techniques.

For illustration purposes, let us assume there are two classifiers $A$ and $B$. $A$ labels data for $B$, and $B$ labels data for $A$ until there are no unlabeled data left or none can be labeled due to a lack of enough confidence in the label. The decision to label data for the other classifier is taken on the basis of statistical techniques. Following Goldman and Zhou (2000) we construct confidence intervals for $A$ and $B$ using 10-fold cross-validation. Let $l_A$, $l_B$, $h_A$, $h_B$, be the lower and upper confidence intervals for $A$ and $B$, respectively. Additionally, the upper and lower confidence intervals for each class $k$ can be defined as $l_{Ak}$, $h_{Ak}$, $l_{Bk}$, and $h_{Bk}$. These intervals define the confidence assigned to each of the classifiers in their labeling mechanism. In addition, the amount of data labeled is subject to a noise control mechanism.

Let us consider the co-training round for $A$ that is labeling data for $B$ (the other will follow similarly). Let $X_L$ be the original labeled training set (which is same for $A$ and $B$), $X_{LB}$ be the data labeled by $A$ for $B$, $w_B$ be the conservative estimate for the mislabeling errors in $X_{LB}$, $m = |X_L \cup X_{LB}|$ be the sample size, and $\eta = w_B/|X_L \cup X_{LB}|$ be the noise rate. The relationship between a sample of size $m$, classification noise rate of $\eta$, and hypothesis error of $\varepsilon$ can be given as

$$m = \frac{k}{\varepsilon^2(1-\eta^2)}, \quad k \text{ is a constant, assumed to be 1}$$

This relationship is used to decide if the amount of additional data labeled can compensate for the increase in the classification noise rate. Thus, based on the expressions for $m$ and $\eta$, the conservative estimate of $1/\varepsilon^2$ for classifier B, $q_B$, can be given as

$$q_B = |X_L \cup X_{LB}| \left(1 - 2\left(\frac{2w_B}{|X_L \cup X_{LB}|}\right)\right)^2$$

Similarly, we can compute the conservative estimate for $1/\varepsilon^2$, $q_k$, for each class $k$ in the labeled set:

$$q_k = |X_L \cup X_{LB} \cup X_{Uk}| \left(1 - \left(\frac{2(w_B + w_k)}{|X_L \cup X_{LB} \cup X_{Uk}|}\right)\right)^2,$$

$$|X_{Uk}| = \text{Examples in } X_U \text{ mapped to class k}$$

$$w_k = (1 - l_k) |X_{Uk}|$$

Thus, for classifier A the labeling process consists of two tests:

(i)     $h_{Ak} > l_B$

(ii)    $q_k > q_B$

The first test ensures that the class $k$ used by classifier $A$ to label data has an accuracy at least as good as the accuracy of classifier $B$. The second test is to help prevent a degradation in performance of classifier $B$ due to the increased noise in labels. The unlabeled examples that satisfy both criteria are labeled by classifier $A$ for classifier $B$ and vice versa. This process is repeated until no more unlabeled examples satisfy the criteria for either classifier. At the end of the co-training procedure (no unlabeled examples are labeled), the classifiers learned on the corresponding final labeled sets are evaluated on the test set.

In our experiments we used two different classification algorithms – Naïve Bayes and C4.5 decision tree. We modified the C4.5 decision tree program to convert leaf frequencies into probabilities using the Laplace correction (Provost & Domingos, 2003). For the final prediction, the probability estimates from both Naïve Bayes and C4.5 are averaged.

## 2.2 ASSEMBLE

The ASSEMBLE algorithm (Bennett et al., 2002) won the NIPS 2001 Unlabeled data competition. The intuition behind this algorithm is that one can build an ensemble that works consistently on the unlabeled data by maximizing the margin in function space of both the labeled and unlabeled data. To allow the same margin to be used for both labeled and unlabeled data, Bennett et al. introduce the concept of a pseudo-class. The pseudo-class of an unlabeled data point is defined as the predicted class by the ensemble for that point. One of the variations of the ASSEMBLE algorithm is based on AdaBoost (Freund & Schapire, 1996) that is adapted to the case of mixture of labeled and unlabeled data. We used the ASSEMBLE.AdaBoost algorithm as used by the authors for the NIPS Challenge.

ASSEMBLE.AdaBoost starts the procedure by assigning pseudo-classes to the instances in the unlabeled set, which allows a supervised learning procedure and algorithm to be applied. The initial pseudo-classes can be assigned by using a 1-nearest-neighbor algorithm, *ASSEMBLE-1NN*, or simply assigning the majority class to each instance in the unlabeled datasets, *ASSEMBLE-Class0*. We treat majority class as class 0 and minority class as class 1 in all our datasets. The size of the training set over boosting iterations is the same as the size of the labeled set. Figure 1 shows the pseudo code of the ASSEMBLE.AdaBoost algorithm.

In the pseudo code, $L$ signifies the labeled set, $U$ signifies the unlabeled set, $l$ (size of the labeled set) is the sample size within each iteration and ß is the weight assigned to labeled or unlabeled data in the distribution $D_0$. The initial misclassification costs are biased towards the labeled data, due to more confidence in the labels. In the subsequent iterations of the algorithm, a, indicates the relative weight given to each type of error – either labeled data or unlabeled data error. In the pseudo-code, $f$ indicates the supervised learning algorithm, which is Naïve Bayes for our experiments. $T$ indicates the number of iterations, and $f_t(x_i) = 1$ if an instance $x_i$ is correctly classified and $f_t(x_i) = -1$, if it is incorrectly classified; $\varepsilon$ is the error computed for each iteration $t$, and $D_t$ is the sampling distribution.

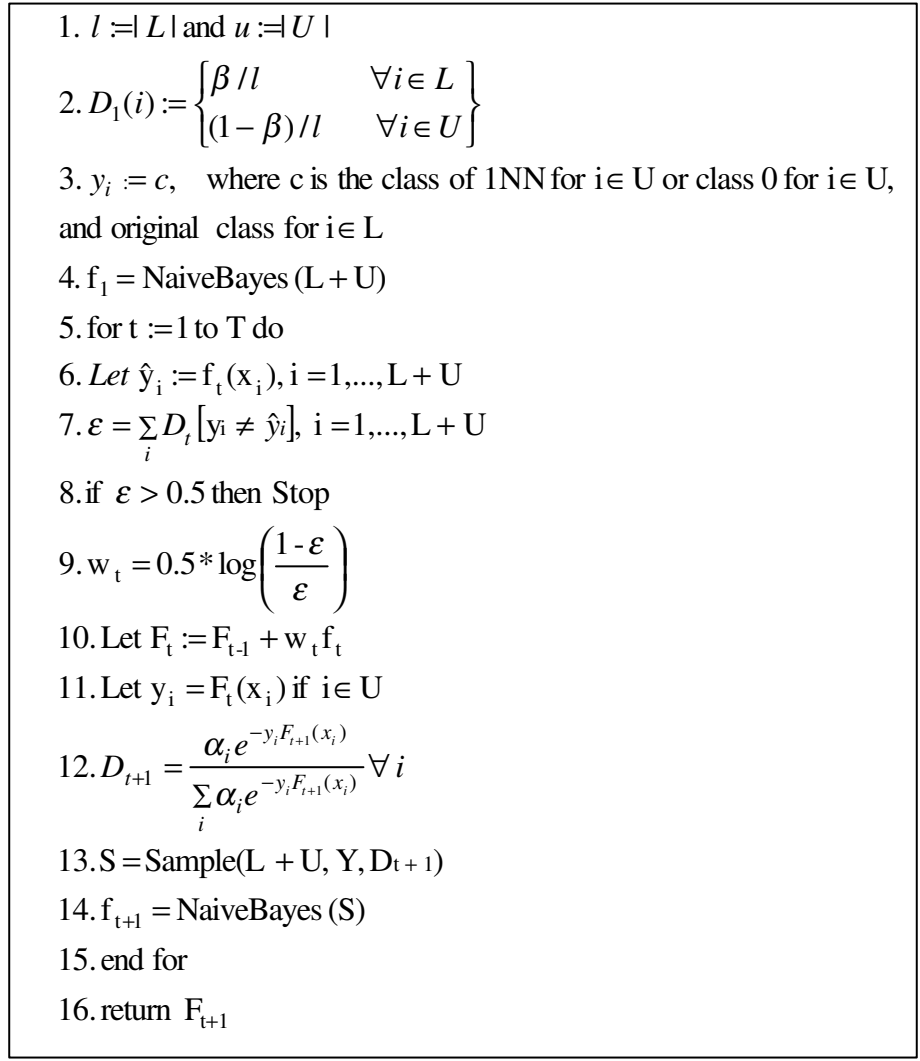

**Figure 1:** ASSEMBLE.AdaBoost pseudo-code

## 2.3 Re-weighting

Re-weighting (Crook & Banasik, 2002) is a popular and simple technique for reject-inferencing in credit scoring, but it has not been considered as a semi-supervised learning technique. It uses information on the examples from approved credit applications, i.e. labeled data, to infer conditional class probabilities for the rejected applications, i.e. unlabeled data.

To apply re-weighting one assumes that the unlabeled data are MAR, i.e.

$$P(y = 1 \mid x, labeled = 1) = P(y = 1 \mid x, labeled = 0) = P(y = 1 \mid x)$$

That is, at any *x*, the distribution of labeled cases having a *y* label is the same as the distribution of "missing" *y* (in the unlabeled population).

There are several variants of re-weighting. According to the one we adopt here, also called extrapolation, the goal is first to estimate the distribution of classes for the labeled data within each score group, and then extrapolate that distribution to the unlabeled data. Thus, a model is learned on the labeled training set and applied to the unlabeled data. The labeled and unlabeled data are grouped by score, i.e. posterior class probability from the model. The probabilities are percentile-binned, and each bin forms a score group. The unlabeled data are assigned labels based on the distribution of classes in corresponding score group from the labeled data. Thus, the key assumption is that the corresponding score-bands in the labeled and unlabeled data have the same distribution for the classes. This can be formulated as:

$$P(y_k \mid S_j, X_L) = P(y_k \mid S_j, X_U)$$

$$\therefore y_{kj}^L / X_{Lj} = y_{kj}^U / X_{Uj}$$

where $S_j$ is the score band of group $j$, $y_{kj}^L$ is the number of labeled examples belonging to class $y_k$ in group $j$, and $y_{kj}^U$ is the proportion of unlabeled examples that could be class $y_k$ in the score group $j$. The number of labeled cases in $j$ are weighted by $(|X_{Lj}+X_{Uj}|)/|X_{Lj}|$, which is the probability sampling weight.

To explain the re-weighting concept, let us consider the example given in Table 1. Let 0 and 1 be the two classes in the datasets. The score group of (0.6 - 0.7) is a bin on the posterior class probabilities of the model. The group weight for the example in Table 1 can be computed as $(X_L+X_U)/X_L = 1.2$. Therefore, weighting the number of *class0* and *class1* in the group, we get *class0* = 12; *class1* = 108. Thus, we label at random 2 (=12-10) data points from the unlabeled set as *class0*, and 18 (=108-90) data points from the unlabeled set as *class1*.

After labeling the unlabeled examples we learn a new model on the expanded training set (inclusive of both labeled and unlabeled data).

| Score Group | Unlabeled | Labeled | Class 0 | Class 1 |
|---|---|---|---|---|
| 0.6 - 0.7 | 20 | 100 | 10 | 90 |

Table 1: Re-weighting example

## 2.4 Common Components Using EM

The idea behind this technique is to eliminate the bias in the estimation of a generative model from labeled/unlabeled data of the MAR type. The reason for this bias is that in the case of such data the distributions of x given y in the labeled and unlabeled sets are different. We can remove this bias by estimating the generative model from both labeled and unlabeled data by modeling the missing labels through a hidden variable within the mixture model framework.

In particular, suppose that the data was generated by M components (clusters), and that these components can be well modeled by the densities $p(x \mid j, \theta_j)$, $j = 1,2,...,M$, with $\theta_j$ denoting the corresponding parameter vector. The feature vectors are then generated according to the density:

$$p(x \mid \theta) = \sum_{j=1}^{M} \pi_j p(x \mid j, \theta_j)$$

where $\pi_j \equiv p(j)$ are the mixing proportions that have to sum to one.

The joint data log-likelihood that incorporates both labeled and unlabeled data takes the form:

$$L(\Theta) = \sum_{(x_i, y_i) \in X_l} \log \sum_{j=1}^{M} \pi_j \, p(x_i \mid j, \theta_j) \, p(y_i \mid x_i, j, \xi_j) + \sum_{x_i \in X_u} \log \sum_{j=1}^{M} \pi_j \, p(x_i \mid j, \theta_j)$$

Note that the likelihood function contains a "supervised" term, which is derived from $X_l$ labeled data, and an "unsupervised" term, which is based on $X_u$ unlabeled data.

Consider applying the following simplistic assumption: the posterior probability of the class label is conditionally independent of the feature vector given the mixture component, i.e. $p(y_i \mid x_i, j, \xi_j) = p(y_i \mid j)$. This type of model makes the assumption that the class conditional densities are a function of a common component (CC) mixture (Ghahramani & Jordan, 1994; Miller & Uyar, 1997). In general, we are interested in applying the CC mixture model to solve a classification problem. Therefore we need to compute the posterior probability of the class label given the observation feature vector. This posterior takes the form:

$$p(y_i \mid x_i, \Theta) = \sum_{j=1}^{M} p(j \mid x_i, \Theta) p(y_i \mid x_i, j, \xi_j) = \sum_{j=1}^{M} \left[ \frac{\pi_j \, p(x_i \mid j, \theta_j)}{\sum_{l=1}^{M} \pi_l \, p(x_i \mid l, \theta_l)} \right] p(y_i \mid j)$$

One could also consider applying a separate component (SC) mixture model, where each class conditional density is modeled by its own mixture components. This model was essentially presented by Nigam et al. (2000) for the case of labeled/unlabeled data in document classification. All of these models, as well as more powerful ones that condition on the input feature vector, can be united under the mixture-of-experts framework (Jacobs et al., 1991). The application of this framework to the semi-supervised setting is further studied by Karakoulas and Salakhutdinov (2003).

### 2.4.1 Model Training

Assume that $p(x \mid j, \theta_j)$ is parameterized by a Gaussian mixture component distribution for continuous data. We define the parameter vector for each component, j, to be $\theta_j = (\mu_j, \Sigma_j)$ for $j = 1, 2, ..., M$; where $\mu_j$ is the mean, and $\Sigma_j$ is the covariance matrix of the $j^{th}$ component. In all of our experiments we constrain the covariance matrix to be diagonal. We also have two more model parameters that have to be estimated: the mixing proportion of the components, $\pi_j = p(j)$, and $\beta_{y_i \mid j} = p(y_i \mid j)$ the posterior class probability.

According to the general mixture model assumptions, there is a hidden variable directly related to how the data are generated from the M mixture components. Due to the labeled/unlabeled nature of the data, there is an additional hidden variable for the missing class labels of the unlabeled data. These hidden variables are introduced into the above log-likelihood $L(\Theta)$.

A general method for maximum likelihood estimation of the model parameters in the presence of hidden variables is the Expectation-Maximization (EM) algorithm (Dempster et al., 1977). The EM algorithm alternates between estimating expectations for the hidden variables (incomplete data) given the current model parameters and refitting the model parameters given the estimated, complete data. We now derive fixed-point EM iterations for updating the model parameters.

For each mixture component, $j$, and each feature vector, $i$, we compute the expected responsibility of that component via one of the following two equations, depending on whether the $i^{th}$ vector belongs to the labeled or unlabeled set (*E-Step*):

$$p(j \mid x_i, y_i, \theta^t) = \frac{\pi_j^t \beta_{y_i \mid j} \, p(x_i \mid j, \theta_j^t)}{\sum_{l=1}^{M} \pi_l^t \beta_{y_i \mid l} \, p(x_i \mid l, \theta_l^t)} \qquad x_i \in X_L$$

$$p(j \mid x_i, \theta^t) = \frac{\pi_j^t \, p(x_i \mid j, \theta_j^t)}{\sum_{l=1}^{M} \pi_l^t \, p(x_i \mid l, \theta_l^t)} \qquad x_i \in X_U$$

Given the above equations the solution for the model parameters takes the form (*M-Step*):

$$\pi_j^{t+1} = \frac{1}{N} \left( \sum_{(x_i, y_i) \in X_l} p(j \mid x_i, y_i, \theta^t) + \sum_{x_i \in X_u} p(j \mid x_i, \theta^t) \right)$$

$$\beta_{y_k \mid j}^{t+1} = \frac{\sum_{x_i \in X_l \cap y_i = k} p(j \mid x_i, y_i, \theta^t)}{\sum_{(x_i, y_i) \in X_l} p(j \mid x_i, y_i, \theta^t)}$$

$$\mu_j^{t+1} = \frac{1}{N \pi_j^{t+1}} \left( \sum_{(x_i, y_i) \in X_l} p(j \mid x_i, y_i, \theta^t) x_i + \sum_{x_i \in X_u} p(j \mid x_i, \theta^t) x_i \right)$$

$$\Sigma_j^{t+1} = \frac{1}{N \pi_j^{t+1}} \left( \sum_{(x_i, y_i) \in X_l} p(j \mid x_i, y_i, \theta^t) S_{ij}^t + \sum_{x_i \in X_u} p(j \mid x_i, \theta^t) S_{ij}^t \right)$$

where we define $N = \mid X_l \mid + \mid X_u \mid$, and $S_{ij} \equiv (x_i - \mu_j^t)(x_i - \mu_j^t)^T$.

Iterating through the E-step and M-step guarantees an increase in the likelihood function. For discrete-valued data, instead of using a mixture of Gaussians, we apply a mixture of multinomials (Kontkanen et al., 1996).

In general, the number of components can be tuned via cross-validation. However, the large number of experiments reported in this work makes the application of such tuning across all experiments impractical, because of the computational overhead it would entail. Furthermore, the purpose of this paper is to provide insight into learning from labeled and unlabeled data with different techniques, rather than to find out which technique performs the best. For this reason we report the results from experiments using two, six, twelve and twenty-four components. We further discuss this point in Section 4.

## 3. Dealing With Sample Selection Bias

If there is a censoring mechanism that rules the assignment of a label to instances in the data then a model learned on only the labeled data will have a sample selection bias. In this case the decision for labeling can be due to unobserved features that create a dependency between assigning a label and the class label itself. Thus the data are of MNAR type. For example, suppose we are interested in assigning relevance to a text corpus, e.g., web pages, and that someone has subjectively decided which documents to label. The estimates for probability of relevance can then be underestimated, as the training set might not be representative of the complete population. This can result in an increase in the estimation bias of the model, thus increasing the error.

We present two techniques, previously proposed in the literature, for dealing with MNAR data.

### 3.1 Bivariate Probit

Heckman (1979) first studied the sample-selection bias for modeling labor supply in the field of Econometrics. Heckman's sample selection model in its canonical form consists of a linear regression model and a binary probit selection model. A probit model is a variant of a regression model such that the latent variable is normally distributed. Heckman's method is a two-step estimation process, where the probit model is the selection equation and the regression model is the observation equation. The selection equation represents the parameters for classification between labeled or unlabeled data – namely its purpose is to explicitly model the censoring mechanism and correct for the bias – while the observation equation represents the actual values (to regress on) in the labeled data. The regression model is computed only for the cases that satisfy the selection equation and have therefore been observed. Based on the probit parameters that correct for the selection bias, a linear regression model is developed only for the labeled cases.

The following equations present the regression and probit models. The probit model is estimated for $y_2$. The dependent variable $y_2$ represents whether data is labeled or not; $y_2 = 1$ then data is labeled, $y_2 = 0$ then data is unlabeled. The equation for $y_1$, the variable of interest, is the regression equation. The dependent variable $y_1$ is the known label or value for the labeled data. The latent variables $u_1$ and $u_2$ are assumed to be bivariate and normally distributed with correlation $\rho$. This is the observation equation for all the selected cases (the cases that satisfy $y_2 > 0$). Let us denote by $y_1^*$ the observed values of $y_1$. Then we have:

$$y_1 = \beta' x_1 + u_1$$
$$y_2 = \gamma' x_2 + u_2$$
$$u_1, u_2 \approx N[0, 0, \sigma_{u_1}^2, \rho]$$
$$y_1 = y_1^*, if \ y_2 > 0$$
$$y_1 \equiv missing \ \ if \ y_2 \leq 0$$

The estimate of $\beta$ will be unbiased if the latent variables $u_1$ and $u_2$ are uncorrelated, but that is not the case as the data are MNAR. Not taking this into account can bias the results. The bias in regression depends on the sign of $\rho$ (biased upwards if $\rho$ is positive, and downwards if $\rho$ is negative). The amount of bias depends on the magnitude of $\rho$ and $\sigma$. The crux of Heckman's method is to estimate the conditional expectation of $u_1$ given $u_2$ and use it as an additional variable in the regression equation for $y_1$. Let us denote the univariate normal density and CDF of $u_1$ by $\phi$ and $\Phi$, respectively. Then,

$$E[y_1^* \mid x_1, y_2 > 0] = \beta' x_1 + \rho \sigma_{u_1} \lambda$$

$$\text{where } \lambda = \frac{\phi(x_2 \gamma')}{\Phi(x_2 \gamma')}$$

$\lambda$ is also called the *Inverse Mills Ratio* (IMR), and can be calculated from the parameter estimates.

Heckman's canonical form requires that one of the models be regression based. However, in this paper we deal with semi-supervised classification problems. For such problems a class is assigned to the labeled data, $y_1 = 0$ or 1, only if $y_2 = 1$. Given the two equations in Heckman's method, the labeled data are only observed for $y_2 = 1$. Thus, we are interested in $E[y_1|x, y_2 = 1]$ or $P(y_1|x, y_2 = 1)$. For example, to predict whether someone will default on a loan (bad customer), a model needs to be learnt from the accepted cases. Thus, in such scenarios one can use the *bivariate probit* model for two outcomes, default and accept (Crook & Banasik, 2002; Greene, 1998). The bivariate probit model can be written as follows:

$$y_1 = \beta' x_1 + u_1$$
$$y_2 = \beta' x_2 + u_2$$
$$y_1 = y_1^* \text{ for } y_2 = 1$$
$$Cov(u_1, u_2) = \rho$$

If $\rho = 0$ then there is no sample-selection bias associated with the datasets, that is the latent variables are not correlated and using the single probit model is sufficient. The log-likelihood of the bivariate model can be written as (where $\Phi_2$ is the bivariate normal CDF and $\Phi$ is the univariate normal CDF).

$$Log(L) = \sum_{y_2=1, y_1=1} \log F_2[\beta_1 x_1, \beta_2 x_2, \rho]$$
$$+ \sum_{y_2=1, y_1=1} \log F_2[-\beta_1 x_1, \beta_2 x_2, -\rho]$$
$$- \sum_{y_2=0} \log F[-\beta_2 x_2]$$

We apply the bivariate probit model in the semi-supervised learning framework by introducing a selection bias in various datasets.

## 3.2 Sample-Select

Zadrozny and Elkan (2000) applied a sample-selection model to the KDDCup-98 donors' dataset to assign mailing costs to potential donors. There is sample selection bias when one tries to predict the donation amount from only donors' data. Using Naïve Bayes or the probabilistic version of C4.5, they computed probabilistic estimates for membership of a person in the "donate" or "not donate" category, i.e. *P(labeled=1|x)*. They then imputed the membership probability as an additional feature for a linear regression model that predicts the amount of donation. Zadrozny and Elkan did not cast the above problem as a semi-supervised learning one. We adapted their method for the semi-supervised classification setting as follows:

$$x_{n+1} = P(labeled = 1 \mid x, L \cup U)$$

$$\therefore P(y = 1 \mid x \cup x_{n+1}, L)$$

*L* is the labeled training set, *U* is the unlabeled set, *labeled=1* denotes an instance being labeled, *labeled = 0* denotes an instance being unlabeled, and *y* is the label assigned to an instance in the labeled set. A new training set of labeled and unlabeled datasets ($L \cup U$) is constructed, wherein a class label of 1 is assigned to labeled data and a class label of 0 is assigned to unlabeled data. Thus, a probabilistic model is learned to classify data as labeled and unlabeled data. The probabilities, *P(labeled = 1|x),* assigned to labeled data are then included as another feature in the labeled data only. Then, a classifier is learned on the modified labeled data (with the additional feature). This classifier learns on "actual" labels existing in the labeled data.

For our experiments, we learn a Naïve Bayes classifier on *L $\cup$ U,* and then impute the posterior probabilities as another feature of *L*, and relearn the Naïve Bayes classifier from *L*. For the rest of the paper we will call this method *Sample-Select.*

While we apply the Sample-Select method along with other semi-supervised techniques, we only apply the bivariate probit model to the datasets particularly constructed with the sample-selection bias. This is because Sample-Select is more appropriate for the MAR case rather than the MNAR case due to the limiting assumption it makes. More specifically, in the MNAR case we have

$$P(labeled, y|x, \xi, \omega) = P(y|x, labeled, \xi) * P(labeled|x, \omega)$$

However, since Sample-Select assumes

$$P(y|x, labeled=0, \xi) = P(y|x, labeled=1, \xi)$$

this reduces the problem to MAR.

## 4. Experimental Set-up

The goal of our experiments is to investigate the following questions using the aforementioned techniques on a variety of datasets:

(i)     What is the effect of independence or dependence among features?

(ii)    How much of unlabeled vs. labeled data can help?

(iii)   What is the effect of label noise on semi-supervised techniques?

(iv)   What is the effect of sample-selection bias on semi-supervised techniques?

(v)    Does semi-supervised learning always provide an improvement over supervised learning?

For each of these questions we designed a series of experiments real-world and artificial datasets. The latter datasets were used for providing insights on the effects of unlabeled data within a controlled experimental framework. Since most of the datasets did not come with a fixed labeled/unlabeled split, we randomly divided them into ten such splits and built ten models for each technique. We reported results on the average performance over the ten models for each technique. Details for this splitting are given in Section 4.1. Each of the datasets came with a pre-defined test set. We used those test sets to evaluate performance of each model. The definition of the AUC performance measure is given in Section 4.2.

As mentioned earlier, the purpose of this work is to provide insight into learning from labeled and unlabeled data with different techniques, rather than to find out which technique performs the best. The latter would have required fine-tuning of each technique. However, to understand the sensitivity of the various techniques with respect to key parameters, we run experiments with different parameter settings. Thus, for the common-component technique, we tried 2, 6, 12, and 24 components; for ASSEMBLE-1NN and ASSEMBLE-Class0 we tried 0.4, 0.7 and 1.0 as values for α, the parameter that controls the misclassification cost; for co-training we tried 90%, 95% and 99% confidence intervals for deciding when to label. Appendix A (Tables 6 to 9) presents the results for these four techniques, respectively. The common-component technique with 6 components gave consistently good performance across all datasets. We found marginal or no difference in the performance of co-training for different confidence intervals. We found small sensitivity to the value of a at lower amounts of labeled data, but as the labeled data increases the differences between the values of a for various datasets becomes smaller. Moreover, not all datasets exhibited sensitivity to the value of a. Our observations agree with Bennett et al. (2002) that choice of a might not be critical in the cost function.

Thus, for analyzing the above questions we set the parameters of the techniques as follows: 6 components for common-component mixture, α = 1 for ASSEMBLE-1NN and ASSEMBLE-Class0, and confidence of 95% for co-training.

## 4.1 Datasets

We evaluated the methods discussed in Sections 2 and 3 on various datasets, artificial and real-world, in order to answer the above questions. Table 2 summarizes our datasets. All our datasets have binary classes and are unbalanced in their class distributions. This feature makes the datasets relevant to most of the real-world applications as well as more difficult for the semi-supervised techniques, since these techniques can have an inductive bias towards the majority class. To study the effects of feature relevance and noise we constructed 5 artificial datasets with different feature characteristics and presence of noise. We modified the artificial data generation code provided by Isabelle Guyon for the NIPS 2001 Workshop on Variable and Feature Selection (Guyon, 2001). The convention used for the datasets is: A_B_C_D, where

- A indicates the percentage of independent features,
- B indicates the percentage of relevant independent features,
- C indicates the percentage of noise in the datasets, and
- D indicates the percentage of the positive (or minority) class in the datasets.

The independent features in the artificial datasets are drawn by N(0,1). Random noise is added to all the features by N(0,0.1). The features are then rescaled and shifted randomly. The relevant features are centered and rescaled to a standard deviation of 1. The class labels are then assigned according to a linear classification from a random weight vector, drawn from N(0,1), using only the useful features, centered about their mean. The mislabeling noise, if specified, is added by randomly changing the labels in the datasets. The naming of the artificial datasets in Table 2 are self-explanatory using the convention outlined earlier.

| Datasets | Training Size \|L+U\| | Testing Size | Class-distribution (majority:minority) | Features |
|---|---|---|---|---|
| 30_30_00_05 | 8000 | 4000 | 95:5 | 30 |
| 30_80_00_05 | 8000 | 4000 | 95:5 | 30 |
| 80_30_00_05 | 8000 | 4000 | 95:5 | 30 |
| 80_80_00_05 | 8000 | 4000 | 95:5 | 30 |
| 30_80_05_05 | 8000 | 4000 | 95:5 | 30 |
| 30_80_10_05 | 8000 | 4000 | 95:5 | 30 |
| 30_80_20_05 | 8000 | 4000 | 95:5 | 30 |
| SATIMAGE (UCI) | 4435 | 2000 | 90.64:9.36 | 36 |
| WAVEFORM (UCI) | 3330 | 1666 | 67:33 | 21 |
| ADULT (UCI) | 32560 | 16281 | 76:24 | 14 |
| NEURON (NIPS) | $530^{L}$+ $2130^{U}$ | 2710 | 71.7:28.3 | 12 |
| HORSE-SHOE (NIPS) | $400^{L}$+ $400^{U}$ | 800 | 61.75:38.25 | 25 |
| KDDCUP-98 (UCI) | $4843^{L}$+ $90569^{U}$ | 96367 | 0.6:99.4 | 200 |

Table 2: Datasets details

Three of our real-world datasets – waveform, satimage, and adult – are from the UCI repository (Blake et al., 1999). We transformed the original six-class satimage datasets into a binary class problem by taking class 4 as the goal class (y=1), since it is the least prevalent one (9.73%), and merging the remaining classes into a single one. Two of the datasets come from the NIPS 2001 (Kremer & Stacey, 2001) competition for semi-supervised learning with pre-defined labeled/unlabeled subsets.

We randomly divided ten times all the above artificial and UCI datasets into five different labeled-unlabeled partitions: (1,99)%, (10,90)%, (33,67)%, (67,33)%, (90,10)%. This provided us with a diverse test bed of varying labeled and unlabeled amounts. We report the results averaged over the ten random runs.

The last real-world dataset is from the KDDCup-98 Cup (Heittich & Bay, 1999; Zadrozny & Elkan, 2000). This is a typical case of sample selection bias, i.e. MNAR. We converted the original KDDCup-98 donors regression problem into a classification problem so that is consistent with the semi-supervised classification setting of the experiments. We considered all the respondents to the donation mailing campaign as the labeled set and the non-respondents as the unlabeled set. We then transformed the labeled set into a classification problem, by assigning a label of "1" to all the individuals that make a donation of greater than \$2, and a label of "0" to all the individuals that make a donation of less than \$2. This gave us a very difficult dataset since due to the sample selection bias built into the dataset the labeled and unlabeled sets had markedly different class distributions. The positive class in the training set is 99.5%, while in the testing set it is only 5%.

We also created a biased version of the Adult dataset and the (30_80_00_05) artificial dataset with MAR labels in order to study the effect of this missing label mechanism on the techniques. More specifically, for the biased Adult dataset we constructed the labeled and unlabeled partitions of the training set by conditioning on the education of an individual. Using this censoring mechanism all the individuals without any post high-school education were put into the unlabeled set, and the rest into the labeled set. Thus, we converted the original classification problem into a problem of return of wages from adults with post high-school education. For the (30_80_00_05) dataset we divided it into labeled and unlabeled sets by conditioning on two features such that: if $(x_{in} <= c_i \,||\, x_{jn} <= c_j)$, $n \in U$. This introduced a selection bias into the construction of the labeled and unlabeled sets. In both datasets the test set was not changed. We further discuss the above three biased datasets in Section 5.3.

## 4.2 Learning And Performance Measurement

In most of our experiments we use the Naïve Bayes algorithm as the base learner. For the co-training experiments we use a C4.5 decision tree learner in addition to Naïve Bayes. To apply the Naïve Bayes algorithm, we pre-discretize the continuous features using Fayyad and Irani's entropy heuristic (Fayyad & Irani, 1993).

We use Area Under the Receiver Operating Characteristic (ROC) curve (AUC) as the performance metric in our experiments (Hand, 1997). Due to the imbalanced class distribution of the datasets studied in this paper we chose to use AUC over the standard classification accuracy metric, because the classification accuracy can be misleading for unbalanced datasets. To compare the performance of a model with that of a random model we define AUC as

$$AUC := 2 * auc - 1$$

where $auc$ is the original area and $AUC$ is the normalized one. Thus, with this formulation, AUC is normalized with respect to the diagonal line in the ROC space that represents the random performance. If the ROC curve of a model is below the diagonal line, then AUC is worse than random and hence negative.

## 5. Empirical Analysis

In the following we present the analysis of the results from our experiments. We structure the analysis around the questions stated in Section 4. Figures 2 to 6 show the AUCs for the various datasets. The Figures show the performance of each of the semi-supervised technique, and the (supervised) Naïve Bayes classifier. The x-axis shows the percentage of labeled/unlabeled data in a dataset as (Labeled, Unlabeled)%, and the y-axis shows the AUC from applying the semi-supervised and supervised techniques. For each of the datasets that does not come with a pre-defined labeled/unlabeled split we average the AUC over 10 random (labeled/unlabeled split) runs. We present the results in charts using the following convention and order for the techniques:

- *Supervised*: Supervised Naïve Bayes
- *ASS1NN*: ASSEMBLE-1NN
- *ASSCLS0*: ASSEMBLE-Class0
- *Samp-Sel*: Sample-Select
- *Reweight*: Reweighting
- *Co-train*: Co-training
- *CC*: Common-Component Mixture

### 5.1 What Is The Effect Of Independence Or Dependence Among Features?

We first investigate the effect of feature independence on semi-supervised learning. For this purpose we use artificial datasets. We consider the first four artificial datasets from Table 2 that do not have noise. Since these datasets were constructed with varying amounts of feature independence and relevance, they offer a diverse test-bed. The results are shown in Figure 2. The results are averaged for 10 different runs for each of the (labeled, unlabeled)% splits.

The two graphs on the left in Figure 2 show the effect of increasing the amount of feature independence, whereas the two graphs on the right show the effect of increasing the amount of feature relevance amongst the independent features. It is worth pointing out that when the amount of feature independence is increased without increasing the amount of feature relevance (top-left to bottom-left) adding unlabeled data hurts performance. When the amount of independence and relevance increases (bottom-right graph) then more semi-supervised learning techniques improve performance over the supervised learner for different amounts of labeled/unlabeled data. This is because in that case the Naive Bayes model underlying the semi-supervised techniques is a better approximation of the true generative model.

Some of the semi-supervised techniques were either comparable or better than learning a classifier from the labeled datasets. The largest improvement in the average AUC was observed for the (1,99)% case across all four combinations of feature independence and relevance, as one would expect. However, (some of) the results were not statistically significant as for some of the (random) labeled and unlabeled splits, the labeled set contained only one or two examples of class 1, thus making it a highly skewed training set. In those cases, the performance was worse than random. This led to a high variance in AUC, causing the confidence intervals to overlap. The lowest gain observed from using semi-supervised learning techniques over learning a supervised Naïve Bayes classifier is in the case of the artificial datasets with only 30% independent and 30% relevant features. This is where the Naive Bayes model achieves relatively high AUC even with only $1/3^{rd}$ of he instances in the labeled set, and addition of more unlabeled data does not help much.

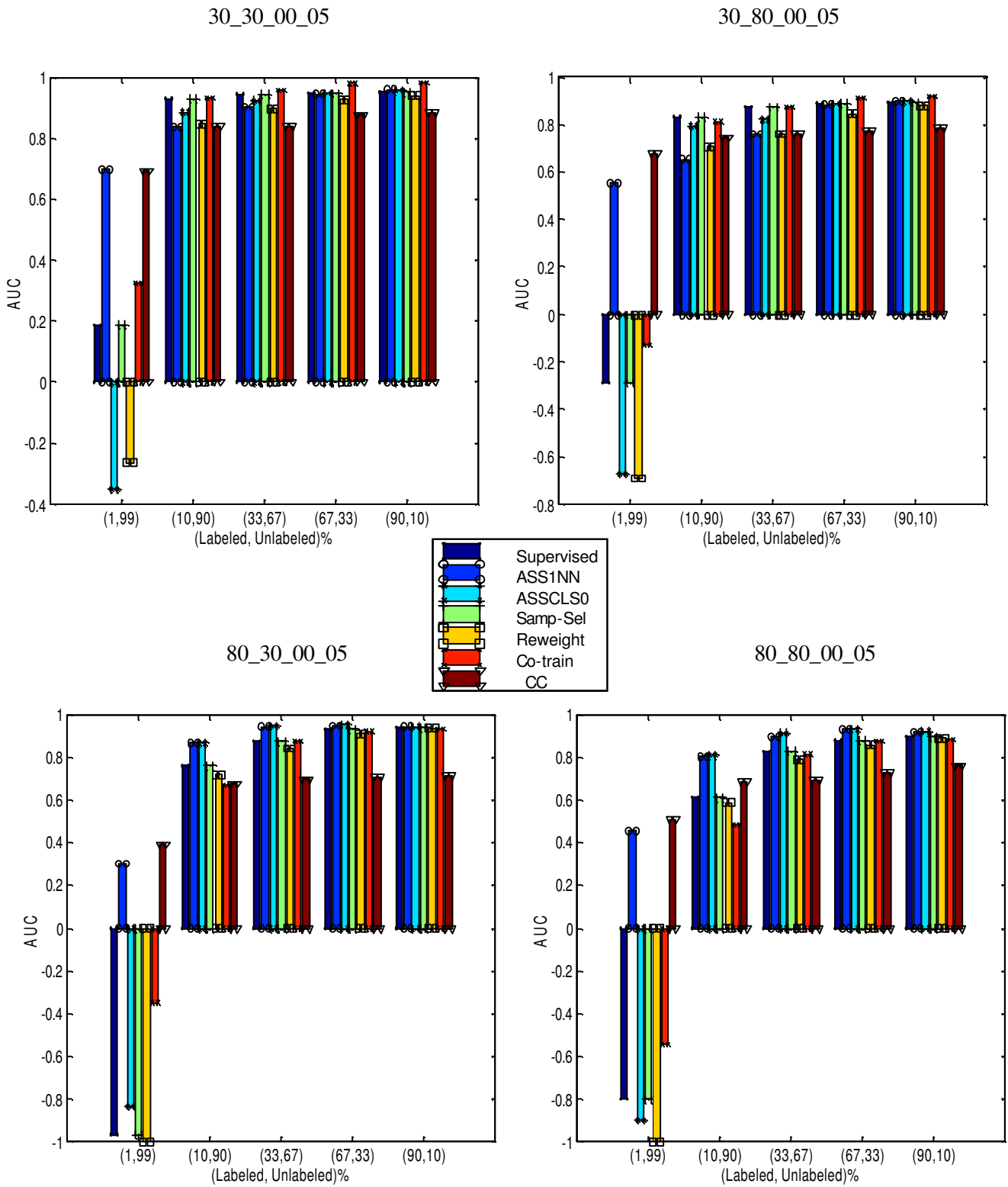

Figure 2 : Artificial datasets without noise. Effect of different amounts of feature independence (top to bottom) and relevance (left to right) at different percentages of labeled and unlabeled data.

Regarding specific techniques, re-weighting is consistently lower than learning a Naïve Bayes classifier on the labeled set only. In re-weighting, the labels are assigned to the unlabeled data as a function of $P(y|x)$. With smaller amounts of labeled data the function could be overfitting given the rich feature space. It is also evident from the much lower performance of the Naive Bayes model on the testing set. Sample-Select does not perform better than the supervised learner even in the $(1,99)\%$ case.

At higher amounts of independence and relevance, ASSEMBLE-1NN and ASSEMBLE-Class0 are (almost) always better than the supervised technique. Co-training is also sensitive to the amount of feature relevance and independence. At higher amounts of independence and relevance, co-training becomes more sensitive to the amount of labeled and unlabeled data in the mixture.

The common-component mixture approach provides a large improvement in the average AUC across all experiments when the amount of labeled data is very small, i.e. $(1,99)\%$, as one would expect. The common-component mixture model seems to be more sensitive to the amount of independent features (top vs. bottom graphs in Figure 2) than the other semi-supervised techniques considered in this paper. This is probably because features that are not correlated with each other might not be very useful for clustering. The goal of clustering is to group feature vectors that are "cohesive" and "distinct" (Fisher, 1987). Thus, correlated features are likely to be more relevant for clustering than independent features. In fact, Talavera (2000) proposes a feature-dependency measure based on Fisher's results for selecting features for clustering. Thus, the decrease in the performance of the common-component mixture technique (top vs. bottom graphs in Figure 2) can be attributed to the increase in the number of independent features.

## 5.2 How Much of Unlabeled vs. Labeled Data Can Help?

To discuss the trade-off between labeled and unlabeled data, we present the results on the UCI and NIPS competition datasets. Figure 3 shows the result on the three UCI datasets – satimage, waveform, and adult. It is evident from the graphs that, as with the artificial datasets, the maximal gain of semi-supervised learning over supervised learning occurs at smaller amounts of labeled data. At lower amounts of labeled data, the common-mixture model gives the biggest gain over Naive Bayes amongst all semi-supervised learning techniques. Also, common component was relatively stable in its performance as the amount of labeled data is increased. We note that the waveform dataset is helped by unlabeled data even at the $(90,10)\%$ split, when using ASSEMBLE-1NN, ASSEMBLE-Class0 and common component. For the adult dataset, most of the semi-supervised techniques do not help for any amount of labeled/unlabeled split. In general, addition of labeled data helped co-training, re-weighting, and Sample-Select. Co-training noted the largest improvement compared to other techniques as we increased to the $(90,10)\%$ split, namely to more labeled data.

Figure 4 shows the results on the two NIPS datasets – neuron and horseshoe. For the neuron dataset, ASSEMBLE-1NN provides a marginal improvement over Naïve Bayes. Co-training is the only technique that helps on the horseshoe dataset. It is worth pointing out that horseshoe has a $(50, 50)\%$ split in labeled and unlabeled data.

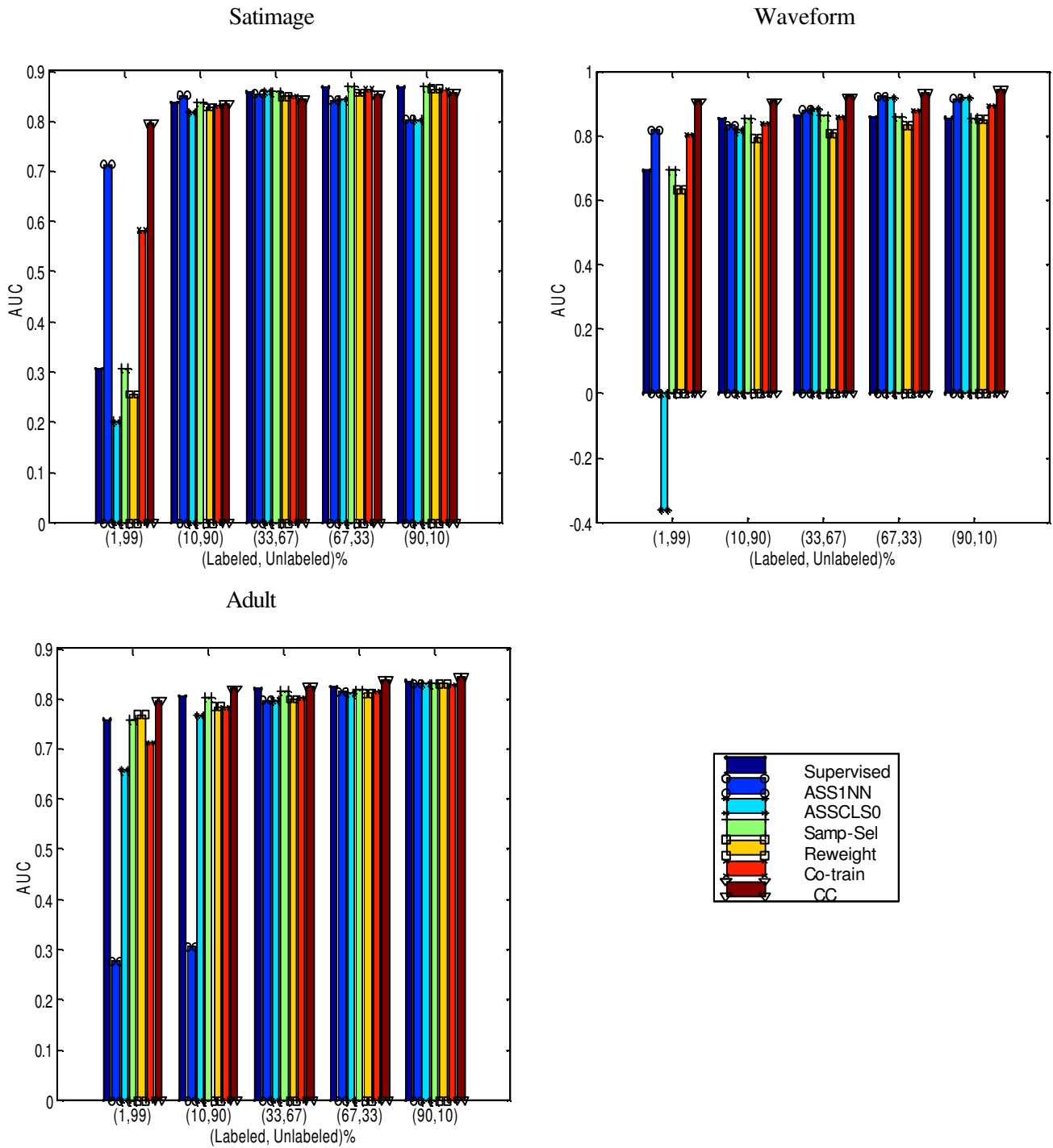

Figure 3: Effect of different percentages of labeled and unlabeled data on three UCI datasets.

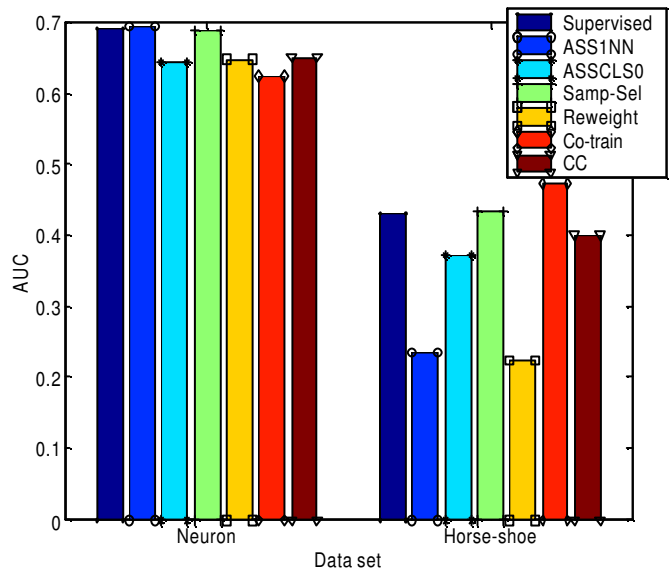

Figure 4: Performance comparison on the two datasets from the NIPS semi-supervised learning competition.

Thus, the amount of labeled and unlabeled data is very much domain dependent. No consistent observation can be drawn from the above experiments. Nevertheless, on average some of the semi-supervised techniques – common-component and ASSEMBLE-1NN – show improvement over the supervised learner for most of the datasets, particularly at the (1,99)% split.

Most of the prior work states that unlabeled data will only help if there is very small amount of labeled data (Nigam et al., 2000; Cozman et al., 2003). However, we observe that even if there is very small amount of labeled data, depending on the technique unlabeled data might not help (see adult in Figure 3). And even if there is a large amount of labeled data unlabeled data can help (see waveform in Figure 3).

### 5.3 What Is The Effect Of Label Noise On Semi-supervised Techniques?

We considered this question because label noise could be detrimental to learning. Therefore if the initial label estimates from a semi-supervised learning technique based on the labeled data are not accurate then adding unlabeled data might provide minimal gain, if any. To understand the sensitivity of the semi-supervised techniques to the amount of noise, we added 5%, 10%, and 20% mislabeling noise to the 30_80_*_05 artificial datasets, where "*" denotes the level of mislabeling noise.

Figure 5 shows the effect of noise on each of the semi-supervised techniques. As before, unlabeled data help only when the amount of labeled data is relatively very small, i.e. (1,99)%. Once again, the common-component mixture and ASSEMBLE-1NN are the ones exhibiting the biggest improvement over Naive Bayes, particularly at (1,99)%. At (10,90)%, ASSEMBLE-Class0 also improves over Naïve Bayes. This is observed across all noise levels. It is worth pointing out that for the datasets with 20% mislabeling noise adding unlabeled data improves performance even at the (10,90)% labeled/unlabeled mix. This is in contrast to what one would expect since mislabeling noise could have caused the model learned from the labeled data to deviate from the underlying distribution. Hence, it could have eliminated any potential improvement from adding unlabeled data. That improvement is probably because the supervised learner is overfitting in the (1,99)% and (10,90)% datasets with the 20% mislabeling noise, due to the high level of noise and small amount of labeled data. This overfitting does not occur at lower levels of noise. Thus, in those two datasets the unlabeled data are acting as a regularizer by preventing the semi-supervised learner from overfitting.

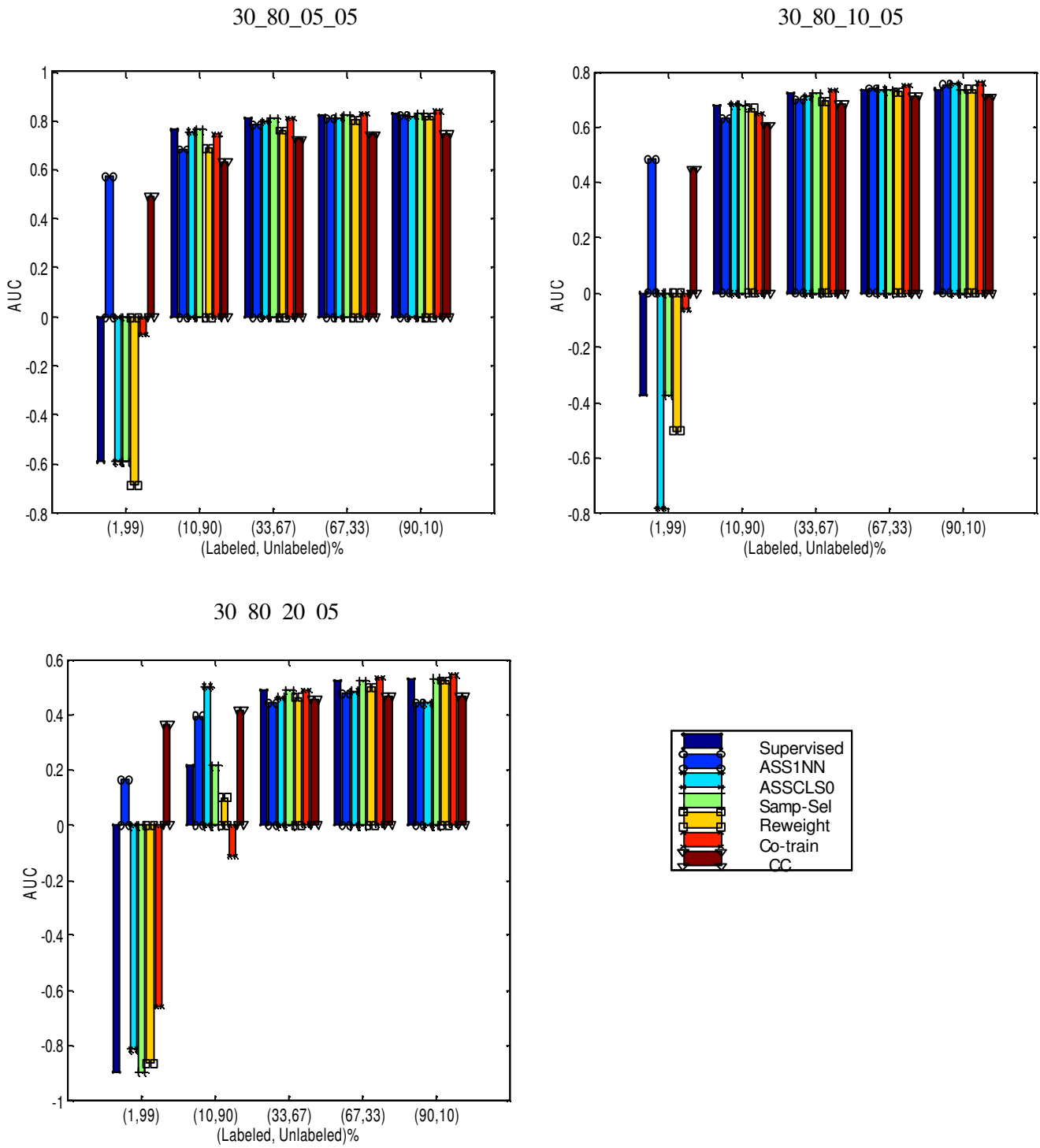

Figure 5: Effect of noise on artificial datasets at varying percentages of labeled and unlabeled data.

### 5.4 What Is The Effect Of Sample-selection Bias On Semi-supervised Techniques?

We used the three datasets with sample selection bias – adult, 30_80_00_05, and KDDCup-98. Section 4.1 detailed the procedure for introducing MAR label bias in the first two datasets. The KDDCup-98 dataset comes with a built-in MNAR bias. Table 3 summarizes the sizes and class distributions of the training set in the three datasets. It is worth noting the different class distributions in the labeled and unlabeled partitions of the training set.

| Datasets | Training Size (Labeled; Unlabeled) | Class Distribution (Labeled; Unlabeled) |
|---|---|---|
| Adult | (17807; 14754) | (66.75, 33.25; 87, 13) |
| 30_80_00_05 | (2033; 5967) | (88.15, 11.85; 97.33, 2.67) |
| KDDCup-98 | (4843; 90569) | (0.5, 99.5; 100, 0) |

Table 3: Composition of the training set in the biased datasets

Table 4 shows the effect of sample selection bias on the feature distributions for the adult and 30_80_00_05 datasets. More specifically, we compared the feature distributions between the labeled and unlabeled sets for each of the two datasets under two different missing data scenarios: biased and random. We used the Kolmogorov-Smirnov statistical test for continuous features and the Chi-squared statistical test for nominal features. The results show that, as expected, for MCAR data there is no statistically significant difference in feature distributions between labeled and unlabeled sets. In contrast there are differences in feature distributions between the two sets when there is some bias in the missing data mechanism.

| | MAR or MNAR data | MCAR data |
|---|---|---|
| Adult | 13 out of 14 different | 0 different |
| 30_80_00_05 | 20 out of 30 different | 0 different |

Table 4: Differences in feature distributions between labeled and unlabeled data under different missing data mechanisms.

Figure 6 shows that the supervised learner (Naïve Bayes) as well as some of the semi-supervised techniques perform at a reasonably good level on the two MAR datasets, but perform very poorly on the MNAR dataset. In particular, co-training and ASSEMBLE-1NN, which did moderately well under the assumption of MCAR, now exhibit significantly worse performance. In fact, ASSEMBLE-Class0 does much better than ASSEMBLE-1NN for the 30_80_00_05 dataset. This is because due to the selection bias the labeled and unlabeled sets have different feature distributions. Thus, the 1NN can be a weaker classification model for initialization of the class labels in ASSEMBLE when the dataset is biased.

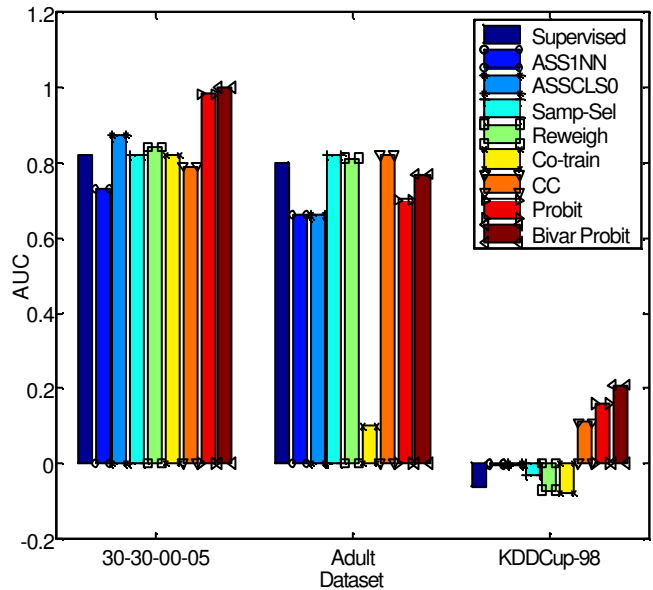

Figure 6: AUC performance for the biased datasets.

Re-weighting does better than a supervised classifier for the adult and 30_80_00_05 MAR datasets and fails on the KDDCup-98 MNAR dataset. This is because re-weighting is not suitable to MNAR datasets since it randomly assigns labels based on the biased posterior probability distribution. Sample-Select has performance comparable to the supervised classifier. Thus, it exhibits poor performance on the KDDCup-98 MNAR dataset. As explained in Section 3.2, Sample-Select is not appropriate for MNAR datasets. The common-component mixture algorithm does not seem to be sensitive to the sample selection bias, even in the MNAR datasets. The common-component technique does better than most of the semi-supervised techniques considered in this paper in the presence of bias. Although the common-component technique scores less than the supervised classifier on the 30_80_00_05 dataset, its performance is consistent with what was observed without a sample selection bias. With the exception of bivariate probit the common-components technique has the best performance amongst the semi-supervised techniques across the three datasets. The supervised Naïve Bayes learner is significantly affected by the introduced selection bias, and its performance decreases from 0.89 to 0.82 at the (approximately) corresponding amounts of labeled and unlabeled data for the 30_80_00_05 dataset. There is a similar effect from the bias in the adult dataset.

Most noteworthy is the improvement provided by the bivariate probit model over the probit model in all three datasets. The bivariate probit successfully corrects for the sample selection bias in the training set. This provides further evidence that just using a biased dataset for prediction on the general population can be detrimental to the classification performance, especially in the case of an MNAR dataset. Part of the reason that probit and bivariate probit perform well on the artificial datasets versus the other techniques is because their assumption of log-normal feature distribution is satisfied in the data. However, they do not perform as well in the Adult dataset since it consists of nominal and real-valued features. In the KDDCup-98 datasets, the bivariate model has the best performance amongst all techniques and the biggest improvement over probit because of the degree of selection bias in the datasets. This is the only dataset that comes with selection bias built in, a true MNAR case. The other two datasets only have a MAR bias.

Based on the results above we believe that in the presence of sample-selection bias availability of more unlabeled data would only help since a better representation of the population could then be induced. Of course, choosing an appropriate learning technique for such data, like bivariate probit or even common-component mixture, is also critical.

## 5.5 Does Semi-Supervised Learning Always Provide An Improvement Over Supervised Learning?

Figure 7 summarizes the performance of each semi-supervised technique across all datasets. The $y$-axis is the average AUC (over the 10 runs) for the datasets that were partitioned into labeled and unlabeled sets 10 random times, and for the other datasets it is the AUC obtained on the available labeled-unlabeled split or the biased labeled-unlabeled split. The bar in the box-plot shows the median performance across all the datasets and labeled-unlabeled partition. As observed in the figure, ASSEMBLE-1NN and common components never do worse than random. Although, the median bar of common component is lower than the median bar of the supervised model, its box is the most compact of all. This highlights the relatively less sensitivity of common components to the size of labeled and unlabeled data. On average, most of the other techniques are usually better than only learning from the labeled data – one can observe this from the distribution of the outliers.

Table 5 contains the statistical significance of the comparisons between the various semi-supervised learning methods and the supervised learning technique (Naïve Bayes). To derive this we grouped the AUCs obtained from the 10 different random runs by each technique across all the datasets. Thus, we had 100 observation points (10 datasets times 10 random runs) for each technique. Since we only had one observation point for the NIPS and biased datasets, we did not include those in this analysis. Using the Kruskal-Wallis statistical test we applied a multiple comparison procedure for comparing the different techniques. Kruskal-Wallis is a non-parametric one-way ANOVA test that does not assume that the values are coming from a normal distribution. The test performs an analysis of variance based on the ranks of the different values and not just their means. We report the Wins-Losses (W-T-L) counts for each semi-supervised learning technique against the supervised learning method. At 1% of labeled data, ASSEMBLE-1NN and common-components mixture have more wins than ties and 1 or 0 loss. The common-component model does not do as well at larger amounts of labeled data, while at smaller amounts of labeled

data – a more prevalent scenario – it seems to help. In general the semi-supervised techniques tie with the supervised one.

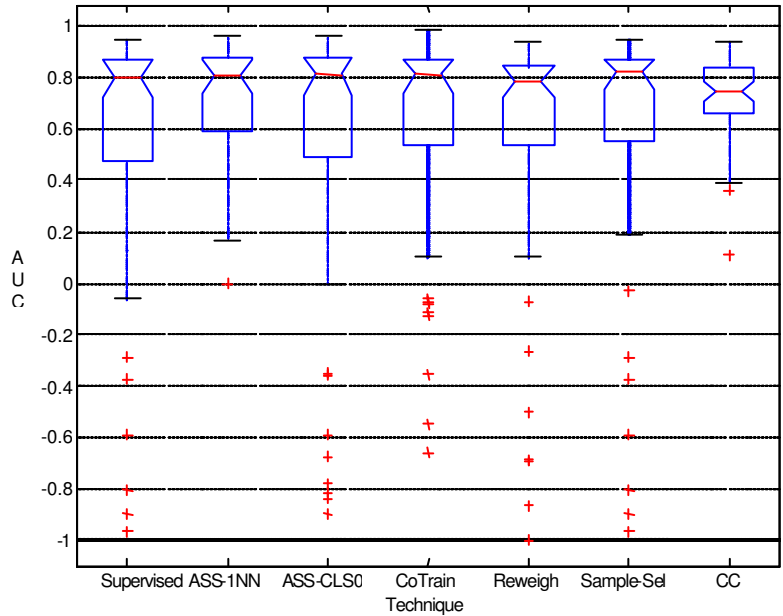

Figure 7: Box plots summarizing the AUC's obtained by various techniques across all datasets. The y-axis indicates the AUC spread, and the x-axis indicates the corresponding method. Each box has lines at the lower quartile, median (red line) and upper quartile. The "+" symbol denotes outliers.

| | (Labeled, Unlabeled)% | | | | |
|---|---|---|---|---|---|
| | (1,99) | (10,90) | (33,67) | (67,33) | (90,10) |
| **Method** | W-T-L | W-T-L | W-T-L | W-T-L | W-T-L |
| Assemble-1NN | 5-4-1 | 1-5-4 | 1-7-2 | 1-7-2 | 2-7-1 |
| Assemble-Class0 | 0-9-1 | 2-7-1 | 2-7-1 | 1-6-3 | 2-6-2 |
| Re-weighting | 0-10-0 | 0-7-3 | 1-4-5 | 0-6-4 | 0-9-1 |
| Sample-Select | 0-10-0 | 0-10-0 | 0-10-0 | 0-10-0 | 1-9-0 |
| Co-training | 0-10-0 | 0-10-0 | 0-10-0 | 2-8-0 | 4-6-0 |
| Common Component | 6-4-0 | 2-4-4 | 1-1-8 | 1-2-7 | 1-4-5 |

Table 5: Win-Tie-Loss (W-T-L) tally for the various techniques versus supervised learning at varying proportions of labeled and unlabeled data.

## 6. Conclusions

We presented various existing techniques from machine learning and econometrics for learning from labeled and unlabeled data. We introduced the concept of sample-selection bias that can be prevalent in real-world labeled sets and potentially detrimental to supervised learning. For this reason we also introduced a bias correction technique from econometrics – bivariate probit.

We empirically evaluated each technique under a set of five main objectives (hypotheses) on a collection of artificial and real-world datasets. For each dataset we varied the proportion of labeled and unlabeled data and compared the performance of each technique against the performance of a supervised classifier. We tried to include datasets from various domains with different characteristics in order to make the study as generalizable as possible. As we observed, different characteristics of data are favorable to different kinds of semi-supervised learning frameworks. Moreover, one may need to perform parameter tuning for optimizing the performance of a specific technique since such optimization was not the focus of this paper.

The main conclusions for each of the five objectives can be enlisted as follows:

- *What is the effect of independence or dependence amongst features?*

We investigated this objective using artificial datasets. We observed that at fewer independent and relevant features (30_*) and at higher amounts of labeled data, the semi-supervised techniques generally do not provide much improvement over the supervised learning method. Given that only 3 or 4 of the 10 independent features are relevant, the fewer labeled examples are usually sufficient to learning the classifier. When the amount of feature independence is increased without increasing the amount of relevant features adding unlabeled data hurts performance. When the amount of independence and relevance increases then more techniques, particularly ASSEMBLE-1NN and ASSEMBLE-Class0, improve performance over the supervised technique. The common-component mixture technique seems to be more sensitive (less effective) when the amount of independent features increases than the other semi-supervised techniques. This is because correlated features are more relevant for clustering.

- *How much of unlabeled vs. labeled data can help?*

For this objective we used both artificial and real-world datasets. We did not observe a consistent pattern that at lower amounts of labeled data, unlabeled data would always help regardless of the technique and domain. We saw that even if there is a small amount of labeled data unlabeled data might not help depending on the technique and domain. Moreover, even if there is a large amount of labeled data, unlabeled data can still help (for example, see Waveform). On average the common-component mixture and ASSEMBLE-1NN helped the most for the (1,99)% labeled/unlabeled mix. Also, common component was relatively stable in its performance as the amount of labeled data is increased. That is, its performance did not deviate much from what it scored at lower amounts of labeled data. As we increased the amount of labeled data, the wins for ASSEMBLE-1NN and common component changed to ties and losses, when compared to the supervised learner. On the contrary, the addition of labeled data helped co-training, re-weighting, and Sample-Select. These techniques ranked very low at the (1,99)% split. Co-training noted the largest improvement compared to other techniques as we increased to (90,10)% split, namely at more labeled data.

- *What is the effect of label noise on semi-supervised learning?*

We investigated this objective using artificial datasets. Noise did not have a detrimental effect on semi-supervised learning. On the contrary, there were fewer semi-supervised models that performed worse than the single supervised model compared to the no-noise cases. This is despite the significant reduction in performance of the single Naive Bayes model. We believe that this is because the single Naive Bayes learner is overfitting as noise level increases. In such cases unlabeled data are acting as a regularizer by preventing the semi-supervised learner to overfit. In fact, at 20% mislabeling noise adding unlabeled data improves performance even at the (10,90)% labeled/unlabeled mix.

- *What is the effect of sample-selection bias on semi-supervised learning?*

For this objective we used two real-world datasets that we transformed them into MAR and a real-world MNAR dataset. Selection bias, especially in the MNAR case, can be detrimental to the performance of the supervised learner and most of the semi-supervised techniques. This is the type of problem where choosing an appropriate learning technique, such as bivariate probit that specifically addresses selection bias or even common-component mixture, can be critical. We observed that ASSEMBLE, which did reasonably well on the datasets for which the labels were missing completely at random, did not perform well on the biased datasets (MAR or MNAR). As expected, Sample-Select did not do well on the MNAR dataset. The one technique that does reasonably well and is again fairly consistent is the common component technique. Bivariate probit has the best performance amongst all techniques in the MNAR dataset. In the presence of selection bias more unlabeled data in conjunction with the appropriate technique seem to always improve performance.

- *Which of the semi-supervised techniques considered is consistently better than supervised learning?*

There is no single technique that is consistently better than the supervised Naïve Bayes classifier for all the labeled/unlabeled splits. At the (1,99)% mix, ASSEMBLE-1NN and common-component mixture were generally better than Naïve Bayes. All the other semi-supervised techniques considered were neither better nor worse than Naïve Bayes. As we increase the amount of labeled data, the wins for ASSEMBLE-1NN and common component move to ties and losses. Common component had more losses as we added more labeled data, particularly in the artificial datasets. Sample-select was comparable to the Naïve Bayes classifier, sometimes slightly better. Re-weighting generally did worse than Naïve Bayes. Co-training noted the largest improvement (over Naïve Bayes) compared to other techniques when we increased the percentage of labeled data.

Semi-supervised learning is still an open problem, especially in the context of MNAR. Current techniques behave very differently depending on the nature of the datasets. Understanding the type of missing data mechanism and data assumptions is key to devising the appropriate semi-supervised learning technique and improving performance by adding unlabeled data. In this work, we dealt with datasets with varying amounts of imbalance in the class distribution. For this purpose, we utilized AUC as the performance measure. An extension to our work could be the utilization of other performance metrics, e.g. classification error, for the analysis of the effects of unlabeled data.

## Acknowledgments

Thanks to Ruslan Salakhutdinov for his help in the experiments. Also, our thanks to Brian Chambers, Danny Roobaert and the anonymous reviewers for their comments and suggestions on earlier versions of this paper.

## Appendix A

Tables 6 to 9 present the results from the experiments with the different parameter settings of the semi-supervised techniques: the number of components in the common-component mixture, the parameter $\alpha$ in ASSEMBLE, and the significance level for the confidence intervals in co-training.

| Dataset | (Labeled, Unlabeled)% | Number of Mixture Components | | | |
|---|---|---|---|---|---|
| | | 2 | 6 | 12 | 24 |
| | | AUC | | | |
| 30_30_00_05 | (1,99) | 0.5488 | **0.6904** | 0.3536 | 0.4529 |
| | (10,90) | 0.6870 | **0.8369** | 0.7245 | 0.7676 |
| | (33,67) | 0.6891 | **0.8377** | 0.7277 | 0.8092 |
| | (67,33) | 0.6911 | **0.8754** | 0.7558 | 0.8057 |
| | (90,10) | 0.6927 | **0.8833** | 0.7458 | 0.8171 |
| 30_80_00_05 | (1,99) | 0.4105 | **0.6753** | 0.2107 | 0.3138 |
| | (10,90) | 0.5145 | **0.7454** | 0.4874 | 0.4834 |
| | (33,67) | 0.5150 | **0.7605** | 0.5344 | 0.5430 |
| | (67,33) | 0.5160 | **0.7760** | 0.5520 | 0.5670 |
| | (90,10) | 0.5167 | **0.7881** | 0.5577 | 0.5831 |
| 80_30_00_05 | (1,99) | 0.2253 | **0.3914** | 0.0982 | 0.1046 |
| | (10,90) | 0.2268 | **0.6728** | 0.1567 | 0.2190 |
| | (33,67) | 0.2317 | **0.6954** | 0.2499 | 0.2674 |
| | (67,33) | 0.2379 | **0.7057** | 0.2516 | 0.3247 |
| | (90,10) | 0.2431 | **0.7138** | 0.2224 | 0.3560 |
| 80_80_00_05 | (1,99) | 0.1323 | **0.5054** | 0.3014 | 0.2713 |
| | (10,90) | 0.2222 | **0.6821** | 0.4553 | 0.4386 |
| | (33,67) | 0.2289 | **0.6921** | 0.4806 | 0.5073 |
| | (67,33) | 0.2392 | **0.7310** | 0.5116 | 0.5302 |
| | (90,10) | 0.2454 | **0.7595** | 0.5099 | 0.5710 |
| Adult | (1,99) | 0.6590 | **0.7954** | 0.6805 | 0.7076 |
| | (10,90) | 0.5736 | **0.8210** | 0.7377 | 0.7453 |
| | (33,67) | 0.7465 | **0.8237** | 0.7523 | 0.7676 |
| | (67,33) | 0.7160 | **0.8378** | 0.7607 | 0.7693 |
| | (90,10) | 0.7578 | **0.8432** | 0.7634 | 0.7807 |
| Satimage | (1,99) | 0.0190 | **0.7972** | 0.6919 | 0.6447 |
| | (10,90) | 0.0264 | **0.8362** | 0.8199 | 0.8299 |
| | (33,67) | 0.0890 | **0.8454** | 0.8259 | 0.8438 |
| | (67,33) | 0.0575 | **0.8544** | 0.8295 | 0.8519 |
| | (90,10) | 0.0832 | **0.8565** | 0.8264 | 0.8520 |
| Waveform | (1,99) | 0.7109 | **0.9074** | 0.8986 | 0.8758 |
| | (10,90) | 0.7135 | 0.9053 | **0.9384** | 0.9298 |
| | (33,67) | 0.7225 | 0.9209 | **0.9394** | 0.9393 |
| | (67,33) | 0.7311 | 0.9334 | 0.9437 | **0.9443** |
| | (90,10) | 0.7348 | 0.9407 | **0.9455** | 0.9449 |
| Horseshoe | | 0.3052 | **0.4000** | 0.3300 | 0.3400 |
| Neuron | | 0.5570 | 0.6500 | 0.6539 | **0.6600** |

Table 6. Sensitivity of common components with respect to number of components for different datasets and labeled/unlabeled split. The bold entries show the components with maximum AUC.

| Dataset | (Labeled, Unlabeled)% | Value of α | | |
|---|---|---|---|---|
| | | 0.4 | 0.7 | 1 |
| | | AUC | | |
| 30_30_00_05 | (1,99) | 0.7244 | **0.7602** | 0.6989 |
| | (10,90) | **0.8621** | 0.8485 | 0.8386 |
| | (33,67) | **0.9295** | 0.9111 | 0.9024 |
| | (67,33) | **0.9557** | 0.9496 | 0.9440 |
| | (90,10) | **0.9599** | 0.9595 | 0.9587 |
| 30_80_00_05 | (1,99) | 0.5538 | **0.5673** | 0.5542 |
| | (10,90) | **0.7047** | 0.6690 | 0.6534 |
| | (33,67) | **0.8502** | 0.8011 | 0.7603 |
| | (67,33) | **0.8989** | 0.8927 | 0.8840 |
| | (90,10) | 0.8992 | **0.9004** | 0.8990 |
| 80_30_00_05 | (1,99) | 0.2730 | **0.3085** | 0.3011 |
| | (10,90) | **0.8824** | 0.8741 | 0.8677 |
| | (33,67) | 0.9305 | 0.9388 | **0.9409** |
| | (67,33) | 0.9416 | 0.9427 | **0.9484** |
| | (90,10) | 0.9409 | **0.9452** | 0.9440 |
| 80_80_00_05 | (1,99) | **0.4760** | 0.4563 | 0.4531 |
| | (10,90) | **0.8610** | 0.8283 | 0.8039 |
| | (33,67) | **0.9046** | 0.9011 | 0.8969 |
| | (67,33) | 0.9207 | 0.9261 | **0.9289** |
| | (90,10) | 0.9127 | **0.9197** | 0.9174 |
| Adult | (1,99) | **0.3059** | 0.2955 | 0.2773 |
| | (10,90) | **0.7754** | 0.7687 | 0.3055 |
| | (33,67) | **0.8027** | 0.7997 | 0.7959 |
| | (67,33) | **0.8210** | 0.8192 | 0.8157 |
| | (90,10) | **0.8323** | 0.8313 | 0.831 |
| Satimage | (1,99) | 0.6940 | 0.6797 | 0.7133 |
| | (10,90) | **0.8566** | 0.8545 | 0.8517 |
| | (33,67) | 0.8582 | **0.8600** | 0.8600 |
| | (67,33) | 0.8402 | 0.8483 | **0.8506** |
| | (90,10) | 0.8165 | 0.8214 | **0.8258** |
| Waveform | (1,99) | 0.8206 | **0.8297** | 0.8160 |
| | (10,90) | **0.8517** | 0.8381 | 0.8327 |
| | (33,67) | **0.8998** | 0.8947 | 0.8815 |
| | (67,33) | 0.9041 | 0.9119 | **0.9137** |
| | (90,10) | 0.8923 | 0.8969 | **0.8989** |
| Horseshoe | | 0.0397 | 0.1778 | **0.2347** |
| Neuron | | **0.7108** | 0.6956 | 0.6929 |

Table 7. Sensitivity of ASSEMBLE-1NN with respect to the value of α for different datasets and labeled/unlabeled split. The bold entries show the α value with maximum AUC.

| Dataset | (Labeled, Unlabeled)% | Value of $\alpha$ | | |
|---|---|---|---|---|
| | | 0.4 | 0.7 | 1 |
| | | AUC | | |
| 30_30_00_05 | (1,99) | **0.2521** | 0.0139 | -0.3505 |
| | (10,90) | **0.9082** | 0.8931 | 0.8849 |
| | (33,67) | **0.9397** | 0.9302 | 0.9235 |
| | (67,33) | **0.9565** | 0.9515 | 0.9473 |
| | (90,10) | **0.9605** | 0.9587 | 0.9578 |
| 30_80_00_05 | (1,99) | **-0.1739** | -0.3721 | -0.6730 |
| | (10,90) | **0.8398** | 0.8105 | 0.7946 |
| | (33,67) | **0.8734** | 0.8466 | 0.8208 |
| | (67,33) | **0.8985** | 0.8947 | 0.8877 |
| | (90,10) | 0.9007 | **0.9025** | 0.8979 |
| 80_30_00_05 | (1,99) | **-0.7405** | -0.6104 | -0.8355 |
| | (10,90) | **0.8874** | 0.8760 | 0.8677 |
| | (33,67) | 0.9334 | 0.9447 | **0.9465** |
| | (67,33) | 0.9410 | 0.9477 | **0.9499** |
| | (90,10) | 0.9412 | 0.9427 | **0.9441** |
| 80_80_00_05 | (1,99) | **-0.6767** | -0.8587 | -0.8990 |
| | (10,90) | **0.8618** | 0.8507 | 0.8096 |
| | (33,67) | 0.9073 | 0.9112 | **0.9112** |
| | (67,33) | 0.9202 | 0.9257 | **0.9290** |
| | (90,10) | 0.9152 | 0.9137 | **0.9201** |
| Adult | (1,99) | **0.6875** | 0.6743 | 0.6573 |
| | (10,90) | **0.7754** | 0.7687 | 0.7671 |
| | (33,67) | **0.8022** | 0.7991 | 0.7975 |
| | (67,33) | **0.8202** | 0.8182 | 0.8148 |
| | (90,10) | **0.8323** | 0.8313 | 0.8306 |
| Satimage | (1,99) | **0.2113** | 0.2024 | 0.2024 |
| | (10,90) | **0.8285** | 0.8255 | 0.8192 |
| | (33,67) | 0.8613 | 0.8619 | **0.8624** |
| | (67,33) | 0.8321 | 0.8455 | **0.8525** |
| | (90,10) | 0.8107 | 0.8162 | **0.8245** |
| Waveform | (1,99) | **0.0017** | -0.2123 | -0.3601 |
| | (10,90) | **0.8424** | 0.8247 | 0.8158 |
| | (33,67) | **0.8902** | 0.8794 | 0.8718 |
| | (67,33) | 0.8945 | 0.9057 | **0.9067** |
| | (90,10) | 0.8884 | 0.8902 | **0.8928** |
| Horseshoe | | 0.2254 | 0.1641 | **0.3711** |
| Neuron | | 0.6572 | **0.6754** | 0.6436 |

Table 8. Sensitivity of ASSEMBLE-Class0 with respect to the value of $\alpha$ for different datasets and labeled/unlabeled split. The bold entries show the $\alpha$ value with maximum AUC.

| Dataset | (Labeled, Unlabeled)% | Confidence Value | | |
|---|---|---|---|---|
| | | 90% | 95% | 99% |
| | | AUC | | |
| 30_30_00_05 | (1,99) | 0.3255 | 0.3255 | 0.3255 |
| | (10,90) | **0.9338** | 0.9343 | 0.9316 |
| | (33,67) | **0.9651** | 0.9571 | 0.9538 |
| | (67,33) | **0.9820** | 0.9799 | 0.9768 |
| | (90,10) | **0.9852** | 0.9851 | 0.9844 |
| 30_80_00_05 | (1,99) | -0.1281 | -0.1281 | -0.1281 |
| | (10,90) | **0.8169** | 0.8129 | 0.8051 |
| | (33,67) | **0.8858** | 0.8748 | 0.8709 |
| | (67,33) | **0.9132** | 0.9121 | 0.9099 |
| | (90,10) | 0.9224 | **0.9227** | 0.9200 |
| 80_30_00_05 | (1,99) | -0.3532 | -0.3532 | -0.3532 |
| | (10,90) | 0.6624 | 0.6694 | **0.7108** |
| | (33,67) | 0.8734 | 0.8734 | 0.8734 |
| | (67,33) | 0.9229 | 0.9229 | 0.9229 |
| | (90,10) | 0.9331 | 0.9331 | 0.9331 |
| 80_80_00_05 | (1,99) | -0.5479 | -0.5479 | -0.5479 |
| | (10,90) | 0.4828 | 0.4828 | 0.4828 |
| | (33,67) | 0.8145 | 0.8145 | 0.8145 |
| | (67,33) | 0.8734 | 0.8734 | 0.8734 |
| | (90,10) | 0.8839 | 0.8839 | 0.8839 |
| Adult | (1,99) | 0.7208 | 0.7208 | 0.7208 |
| | (10,90) | 0.7871 | 0.7871 | 0.7871 |
| | (33,67) | 0.8050 | 0.8050 | 0.8050 |
| | (67,33) | 0.8131 | 0.8131 | 0.8131 |
| | (90,10) | 0.8264 | 0.8264 | 0.8264 |
| Satimage | (1,99) | 0.5213 | **0.5213** | 0.5213 |
| | (10,90) | 0.8308 | **0.8346** | 0.8346 |
| | (33,67) | 0.8566 | 0.8566 | 0.8566 |
| | (67,33) | 0.8625 | 0.8625 | 0.8625 |
| | (90,10) | 0.8650 | 0.8650 | 0.8650 |
| Waveform | (1,99) | 0.8000 | 0.8000 | 0.8000 |
| | (10,90) | **0.8400** | 0.8391 | 0.8377 |
| | (33,67) | 0.8679 | **0.8681** | 0.8681 |
| | (67,33) | 0.8856 | 0.8855 | **0.8859** |
| | (90,10) | **0.8931** | 0.8927 | 0.8930 |
| Horseshoe | | 0.4735 | 0.4735 | 0.4735 |
| Neuron | | 0.6234 | 0.6234 | 0.6234 |

Table 9. Sensitivity of Co-training with respect to the confidence value for different datasets and labeled/unlabeled split. The bold entries show the $\alpha$ value with maximum AUC.